 \documentclass[final,3p,times]{tempcls}

\usepackage{amsmath,amssymb,amsfonts}
\usepackage{amsopn}
\usepackage{graphicx}
\usepackage{subfigure}
\usepackage{multirow}
\usepackage{array}
\usepackage{arydshln}

\journal{}

\begin{document}

\begin{frontmatter}



\title{Collaborative Discriminant Locality Preserving Projections With its Application to Face Recognition}


\author[a,d]{Sheng Huang}
\author[a,b]{Dan Yang\corref{cor1}}
\author[d]{Dong Yang}
\author[d]{Ahmed Elgammal}

\address[a]{College of Computer Science at Chongqing University, Chonqing, 400044, China}
\address[b]{School of Software Engineering at Chongqing University Chonqing, 400044, China}
\address[d]{Department of Computer Science at Rutgers University, Piscataway, NJ, 08854, USA}
\cortext[cor1]{Corresponding author (Dan Yang):dyang@cqu.edu.cn}

\begin{abstract}
We present a novel Discriminant Locality Preserving Projections (DLPP) algorithm named \emph{Collaborative Discriminant Locality Preserving Projection} (CDLPP). In our algorithm, the discriminating power of DLPP are further exploited from two aspects. On the one hand, the global optimum of class scattering is guaranteed via using the between-class scatter matrix to replace the original denominator of DLPP. On the other hand, motivated by collaborative representation, an $L_2$-norm constraint is imposed to the projections to discover the collaborations of dimensions in the sample space. We apply our algorithm to face recognition. Three popular face databases, namely AR, ORL and LFW-A, are employed for evaluating the performance of CDLPP. Extensive experimental results demonstrate that CDLPP significantly improves the discriminating power of DLPP and outperforms the state-of-the-arts.
\end{abstract}


\end{frontmatter}


\section{Introduction}
Subspace learning is a useful technique in computer vision, pattern recognition and machine learning, particularly for solving the dimensionality reduction, feature selection, feature extraction and face recognition tasks. Subspace learning aims to learn a specific subspace of the original sample space, which has some particular desired properties. This topic has been studied for decades and many impressive algorithms have been proposed. The representative subspace learning algorithms include Principle Component Analysis (PCA) \cite{pca,spca,nspca}, Linear Discriminant Analysis (LDA) \cite{lda}, Non-negative Matrix factorization (NMF) \cite{nmf,gnmf}, Independent Component Analysis (ICA) \cite{ica}, Locality Preserving Projections (LPP) and so on. In face recognition, subspace learning is also known as appearance-based face recognition. For example, PCA is known as Eigenfaces, LDA is known as Fisherfaces and LPP is known as Laplacianfaces.

Some recent studies show that the high-dimensional samples may reside on low-dimensional manifolds \cite{lle,isomap} and such manifold structures are essential for data clustering and classification \cite{lap,lpp}. The manifold-based subspace learning algorithms may start from Locality Preserving Projections (LPP). LPP constructs an adjacency matrix to weight the distance between each pair of sample points for learning a projection which can preserve the local manifold structures of data. The weight between two nearby points is much greater than that between two distant points. So if two points are close in the original space, then they will be close in the learned subspace as well. However, the conventional LPP only takes the manifold information into consideration. Many researchers make efforts to improve LPP from different perspectives. Discriminant Locality Preserving Projections (DLPP) \cite{dlpp,tdlppfd,2dlpp} is deemed as one of the most successful extensions of LPP. It improves the discriminating power of LPP via simultaneously maximizing the distance between each two nearby classes and minimizing the original LPP objective. Orthogonal Laplacianfaces (OLPP) \cite{olpp} imposes an orthogonality constraint to LPP to ensure that the learned projections are mutually orthogonal. Parametric Regularized Locality Preserving Projections (PRLPP) \cite{rlpp} regulates the LPP space in a parametric manner and extracts useful discriminant information from the whole feature space rather than a reduced projection subspace of PCA. Furthermore, this parametric regularization can be also employed to other LPP based methods, such as Parametric Regularized DLPP (PRDLPP), Parametric Regularized Orthogonal LPP (PROLPP). Inspired by the idea of LPP, Qiao et al \cite{spp} proposed a novel projection named Sparsity Preserving Projections (SPP) for preserving the sparsity of original sample data and applied it to face recognition.

Our work is mainly based on DLPP. In this paper, we intend to further improve the discriminating power of DLPP from two different aspects. Similar to LPP, DLPP constructs a Laplacian matrix of classes and then improves the discriminating power of LPP via maximizing such matrix. Since the distance between two nearby classes has a greater weight, maximizing the Laplacian matrix of classes actually is equal to maximizing the distance between the nearby classes. Clearly, this strategy cannot guarantee the global optimal class scattering, since the distant classes may be projected closer with each other in such DLPP space than before. In order to obtain the global optimal classes scattering, we use the between-class scatter matrix to replace the Laplacian matrix of classes, which is the denominator of DLPP objective. Moreover, inspired by the idea of the collaborative representation \cite{crc,rcr}, an $L_2$-norm constraint is imposed to the projections, since we believe that not all the dimensions of the samples are equally important and collaboration should exist among the dimensions. For example, if we consider the face images as the samples in face recognition, each dimension of samples is corresponding to a specific pixel in the face images. Clearly, the pixels in the face area of images play a more important role than the pixels in the background area and collaboration exists naturally between the adjacent pixels.

We name the proposed improved DLPP algorithm Collaborative Discriminant Locality Preserving Projections (CDLPP) and apply it to face recognition. Three popular face databases, namely ORL, AR, and LFW-A, are chosen to validate the effectiveness of the proposed algorithm. Extensive experimental results demonstrate that CDLPP remarkably improves the discriminating power of DLPP and outperforms the state-of-the-art subspace learning algorithms with a distinct advantage. Moreover, we also compare CDLPP with four of the most popular face recognition approaches in the recent days, namely Linear Regression Classification (LRC) \cite{lrc}, Sparse Representation Classification (SRC) \cite{sparse}, Collaborative Representation Classification (CRC) \cite{crc} and Relaxed Collaborative Representation Classification (RCR) \cite{rcr}, (These four algorithms are not the subspace learning algorithms).  Even so, CDLPP still outperforms them in all experiments and CDLPP improves the recognition accuracy of RCR from 75\% to 81\% on LFW-A database, which is a very recent challenging face verification and face recognition database. There are three main contributions of our work:
\vspace{-0.2cm}
\begin{enumerate}
 \setlength{\itemsep}{0pt}
 \setlength{\parskip}{0pt}
 \setlength{\parsep}{0pt}
  \item The between-class scatter matrix is used to replace the original denominator of DLPP for guaranteeing the global optimum of class scattering.
  \item According to the fact that collaboration exist among the dimensions of samples, we improve the quality of projections via imposing a collaboration constraint. To the best of our knowledge, our work is the first paper introducing the collaboration of dimensions to the subspace learning. Moreover, this is generalizable to other subspace learning algorithms.
\item A prominent improvement of recognition accuracy of DLPP is obtained by our approach. For example, the gains of CDLPP over DLPP are 12\% and 23\% on the subset 1 and subset 2 of LFW-A database respectively.
\end{enumerate}
\vspace{-0.2cm}

The rest of paper is organized as follows: we introduce related works in section 2; section 3 describes the proposed algorithm; experiments are presented in section 4; the conclusion is finally summarized in section 5.

\section{Related Works}
\label{works}
\subsection{Discriminant Locality Preserving Projections}
Discriminant Locality Preserving Projections (DLPP) \cite{dlpp} is one of the most influential LPP algorithms. It improves the discriminating power of LPP via simultaneously minimizing the original Laplacian matrix of LPP and maximizing the Laplacian matrix of classes. Let $l\times{n}$-dimensional matrix $X=[x_{1},...,x_{n}]$, $x_i\subset\mathcal{R}^{l}$ be the samples and the vector $C=[1,2,...,p]$ be class labels where $p$ is the number of classes. Matrix $X_{c}, c\in{C}$, denotes the samples belonging to class $c$. The $l\times{p}$-dimensional matrix $U=[u_{1},...,u_{i},...,u_{p}]$ denotes the mean matrix where $u_{i}\subset\mathcal{R}^{l}, i\in{C}$ is the mean of the samples belonging to class $i$. A $p$-dimensional row vector $M=w^T{U}=[m_{1},...,m_{i},...,m_{p}], i\in{C}$ presents the projected mean matrix where $l$-dimensional column vector $w$ is a learned projection. Similarly, the projected sample matrix is denoted as a $n$-dimensional row vector $Y=w^T{X}=[y_{1},...,y_{n}]$. DLPP aims to find a set of projections $W=[w_{1},w_{2},...,w_{d}]$ to map the $l$-dimensional original sample space into a $d$-dimensional subspace which can preserve the local geometric structures and scatter classes simultaneously. The $l\times d$-dimensional matrix $W$ denotes projection matrix where $d\ll {l}$.
The original objective of Discriminant Locality Preserving Projections (DLPP) is as follows:
\begin{equation}\label{eqdlpp}
min\frac{{\sum\limits_{c \in C} {\sum\limits_{i,j \in c} {{{({y_i} - {y_j})}^2}} } H_{ij}^c}}{{\sum\limits_{i,j \in c} {{{({u_i} - {u_j})}^2}} {B_{ij}}}}
\end{equation}
where $H_{ij}^c$ and $B_{ij}$ denote the weights of the distance between each two homogenous points and the distance between each two mean points respectively. They are the entries of the respective adjacency weight matrices $H^c$ and $B$. These weights are all determined by the distance between two points (either Cosine distance or Euclidean distance) in the original space. It is not difficult to formulate the numerator and denominator of the objective function in Equation \ref{eqdlpp} into the forms of Laplacian matrices.
\begin{equation}
\hat{w}=\arg\underset{w}{\min}{\frac{w^{T}XLX^{T}{w}}{w^{T}UQU^{T}{w}}} \rightarrow \hat{w}=\arg\underset{w}{\min}{\frac{w^{T}K_{l}{w}}{w^{T}K_{c}{w}}}
\end{equation}
where matrix $L$ is exactly the Laplacian matrix of LPP and matrix $Q$ is the Laplacian matrix of classes.

\subsection{Collaborative Representation Classification}
In recent decade, the sparse representation is very popular and extensive works have emphasized the importance of sparsity for classification \cite{spca,nspca,sparse,dsparse,gsr}. However, some researcher argue that the collaboration actually plays a more important role in the classification rather than the sparsity, since the samples of different classes share similarities and some samples from class $j$ may be very helpful to represent the testing sample with label $i$ \cite{crc,rcr}. In order to collaboratively represent the query sample using $X$ with low computational burden, the $L_2$-norm is used to replace the $L_1$-norm in the objective function of sparse representation. Therefore, the objective function of collaborative representation model is denoted as follows
\begin{equation}\label{}
\hat{p}=\arg\underset{p}{\min}\{||y-X\cdot p||^2_2+\lambda||p||^2_2\}
\end{equation}
where $\lambda$ is the regularization parameter. The role of the regularization term is two-fold. First, it makes the least square solution stable, and second, it introduces a certain amount of \emph{sparsity} to the solution $\hat{p}$ , yet this sparsity is much weaker than that by $L_1$-norm. The solution of this model is
\begin{equation}\label{}
\hat{p}=(X^{T}X+\lambda\cdot I)^{-1}X^Ty
\end{equation}
After we get the solution, we can apply it to classify as the way of sparse representation.

\section{Collaborative Discriminant Locality Preserving Projections }
In the algorithm, we improve discriminating power from two aspects. The first one is to use the between-class scatter matrix to replace the Laplacian matrix of classes, which is the denominator of the objective function of DLPP. The second one is to impose a dimension collaboration constraint to the model.

\begin{figure}[h]
\centering
\subfigure[]{
\centering
\includegraphics[scale=0.32]{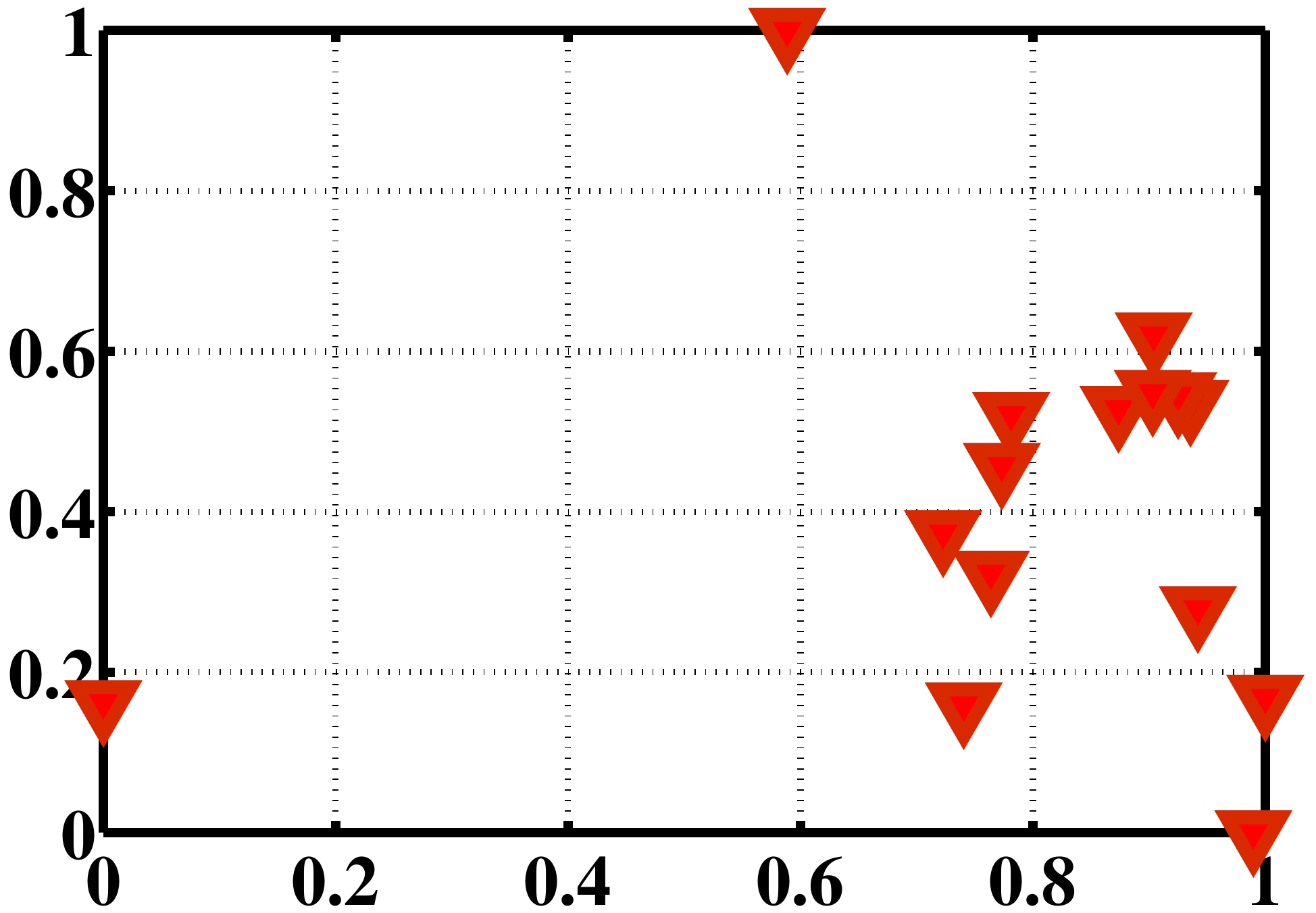}
}
\subfigure[]{
\centering
\includegraphics[scale=0.32]{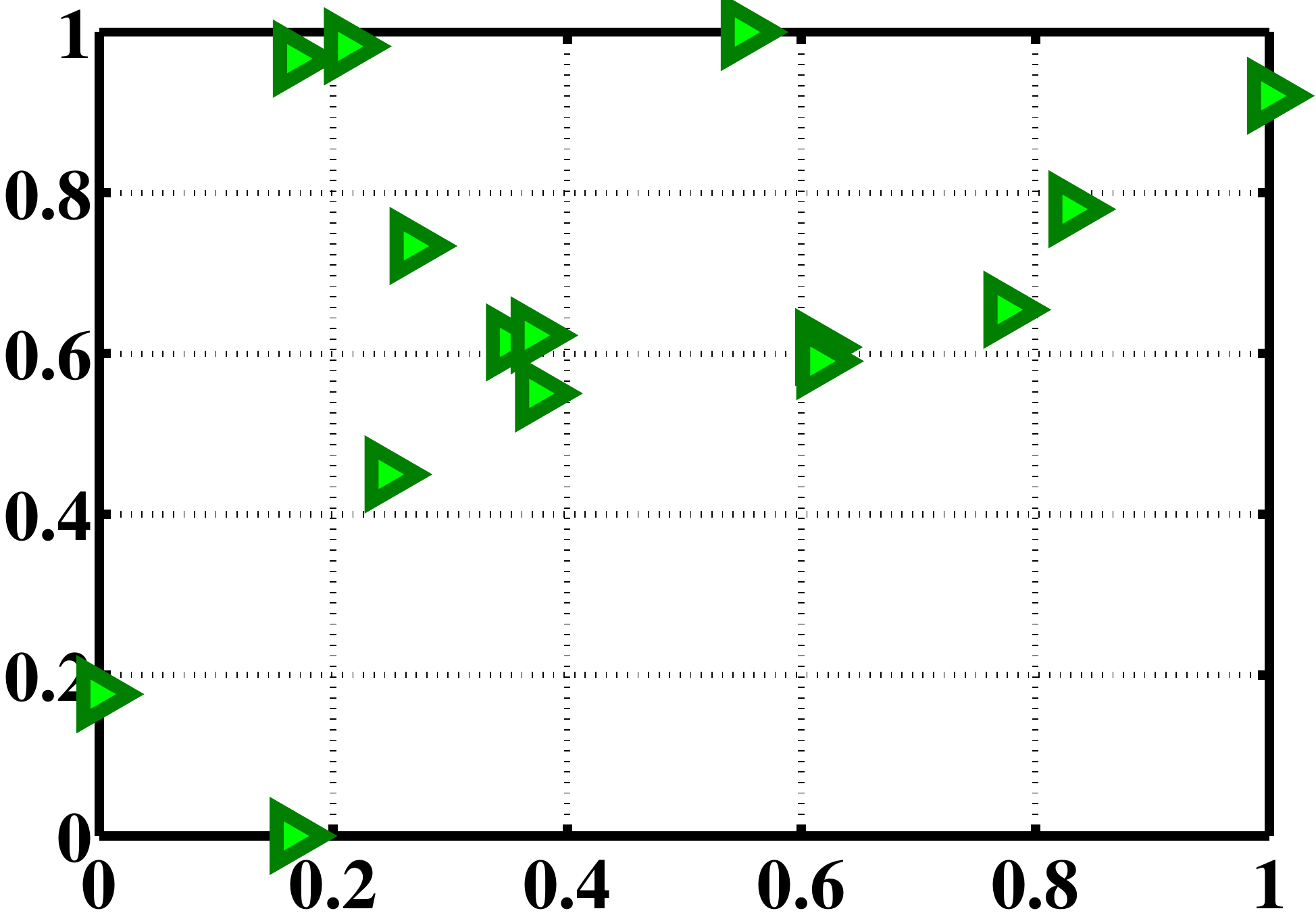}
}
\caption{The illustrations of class scattering abilities of Laplacian matrix of classes and between-class scatter matrix on Yale database \cite{yale} (15 subjects with 11 samples each). \textbf{For clarity, we just draw the center of each class.} (a)The distribution of classes in a subspace learned by maximizing the Laplacian matrix of classes.  (b)The distribution of classes in a subspace learned by maximizing Between-Class Scatter Matrix.}
\label{scattering}
\end{figure}

The core of LPP algorithms is the construction of affinity matrix and the core of the construction of the affinity matrix is the weighting schemes. Several weighting schemes are available for weighting the distance between two samples. The most common used weighting schemes include the dot-product weighting and the heat-kernel weighting \cite{gnmf}. The weighting schemes are all nonlinear and the assigned weight will drop sharply while the distance is increasing. So, the closer points own the greater weight and this strategy makes only the distances between close points able to effectively affect the subspace learning. In that way, if we maximize the denominator of DLPP, which is the Laplacian matrix of classes, only the closer classes can be scattered and the distant classes may be projected much closer. In fisher discriminant analysis, we know that maximizing the between-class scatter matrix can obtain the global optimal classes scatter \cite{lda}. In order to intuitively show the class scattering abilities of between-class scatter matrix and Laplacian matrix of classes, we conduct an experiment via maximizing them respectively on Yale database \cite{yale}. Figure \ref{scattering} shows the results and each point is a class center. The experimental result demonstrates that using the between-class scatter matrix obtains a better performance. Consequently, the discriminating power of DLPP can be further improved via using the between-class scatter matrix instead of the Laplacian matrix of classes, and the new objective function of DLPP is denoted as follows

\begin{equation}\label{eqcslpp}
\hat{w}=\arg\underset{w}{\min}{\frac{w^{T}K_{l}{w}}{w^{T}S_{b}{w}}}
\end{equation}
where
\begin{eqnarray}\label{}\nonumber
\begin{aligned}
w^{T}K_{l}{w}=w^{T}XLX^{T}w={\sum\limits_{c \in C} {\sum\limits_{i,j \in c} {{{({y_i} - {y_j})}^2}} } H_{ij}^c}
\end{aligned}
\end{eqnarray}
and
\begin{eqnarray}\label{}\nonumber
\begin{aligned}
w^{T}S_{b}{w}=w^{T}XC_{p}X^{T}w=\sum\limits_{i \in c} {{n_i}{{({u_i} - \bar u)}^2}}
\end{aligned}
\end{eqnarray}
The $\bar{u}$ is the mean of whole samples, $C_p$ is the $p\times p$ dimensional centering matrix and other notations in this section have been already defined in section \ref{works}. 

We also test the effectiveness of this modification on all the face databases in experiment section. In order to distinguish from the conventional DLPP, we name this modified DLPP \emph{Class Scattering Locality Preserving Projections} (CSLPP). As same as DLPP, the model of CSLPP can be solved by eigenvalue decomposition and the best CSLPP projection $w$ is the eigenvector corresponding to the minimum nonzero eigenvalue of $S_b^{-1}K_{l}$.

As we known, the low-dimensional representation of the sample is achieved by projecting the sample to the learned projection as $y_i=w^Tx_i$. From the perspective of numerical computation, $y_i$ is the sum of dot product of the projection $w$ and the original sample $x_i$, $y_i=x_i\cdot{w}=\sum^l_{j=1}w^jx^j_i$. So, each element of projection, $w^j$, is corresponding to each dimension of the sample, $x_i^j$. A greater value of a specific element of projection, no matter it is positive or negative, has more impact to the $y$. In other words, the dimension of sample corresponding to a greater value element of $w$ should be valued more. Clearly, the role of each dimension of sample is not equally important. For example, if the sample is the face image, and therefore each pixel of image is corresponding to each dimension of sample, the pixels in the face area should play a more important role than the pixels in the background area. Consequently, a good projection $w$ should satisfy the following conditions: the elements of $w$ corresponding to the more important dimensions of sample should own a greater value; the values of elements of $w$ corresponding to the less important dimensions should tend to zero. Moreover, in the subspace learning, the dimensions of sample often highly exceeds the number of samples. So, a good projection $w$ should tend to sparse. Another fact is that the collaboration exists among dimensions in the subspace learning. It can be easily verified in the case of face recognition, for example, the adjacent pixels always collaboratively represent a specific component in the face. The collaboration constraint is imposed to projection $w$ to address above issues, since it can emphasize the importance of the collaborations of dimensions in subspace learning and it is also a relaxed sparsity constraint. According to this optimization, Equation \ref{eqcslpp} can be further modified as follows
\begin{equation}\label{l2scslpp}
\hat{w}=\arg\underset{w}{\min}{\frac{w^{T}K_{l}{w}}{w^{T}S_{b}{w}}}+\beta{\Arrowvert{w}\Arrowvert_2}
\end{equation}
where $\beta>0$ controls the amount of additional collaborations required and we name this new Discriminant Locality Preserving Projections (DLPP) algorithm \emph{Collaborative Discriminant Locality Preserving Projections} (CDLPP). Its objective function can be further formatted as a purely matrix format as follows \footnotemark[1]
\begin{eqnarray}\label{scslppfun}
J(w)=\frac{w^{T}K_{l}{w}}{w^{T}S_{b}{w}}+\beta{\Arrowvert{w}\Arrowvert_2}=\frac{w^{T}K_{l}{w}}{w^{T}S_{b}{w}}+\beta w^T{w}
\end{eqnarray}
Then, the derivative of $w$ can be calculated and let it be equivalent to zero for obtaining the minimum of $J(w)$.
\begin{eqnarray}\label{dscslpp}
\frac{\delta{(J(w))}}{\delta w}&=&\frac{2K_lw-2S_bw{(w^TS_bw)^{-1}w^TK_lw}}{w^TS_bw}+2\beta{w}=0
\end{eqnarray}

Since items $w^{T}K_{l}{w}$ and $w^{T}S_{b}{w}$ can be treated as two unknown scalars and let them be $\alpha$ and $\gamma$ respectively, Equation \ref{dscslpp} can be formulated as follows
\begin{eqnarray}\label{refscslpp}
\frac{2K_lw-2{\gamma^{-1}\alpha}S_bw}{\gamma}+2\beta{w}=0 \nonumber \\
\Rightarrow \frac{\gamma K_lw-{\alpha}S_bw+\gamma^2\beta{w}}{\gamma^2}=0 \nonumber \\
\Rightarrow \qquad \gamma K_{l}w - \alpha S_{b}w +\beta w=0 \\
\Rightarrow \qquad  K_{l}w+\beta I w - \lambda S_{b}w=0  \nonumber \\
\Rightarrow \qquad  S_{b}^{-1}(K_{l}+\beta I)w =\lambda w  \nonumber
\end{eqnarray}
where matrix ${I}$ is an identity matrix and $\lambda=\frac{\alpha}{\gamma}=\frac{w^{T}K_{l}{w}}{w^{T}S_{b}{w}}$ is a scalar. According to Equation \ref{refscslpp}, this problem is also an eigenvalue problem and the best CDLPP projection $w$ is the eigenvector corresponding to the minimum nonzero eigenvalue of $S_{b}^{-1}(K_{l}+\beta I)$.  We can yield the first $d$ CDLPP projections $W=[w_1,...,w_d]$ for face recognition.

\footnotetext[1]{ This formulation can be also transformed as a collaborative representation formulation style format as follows:
\begin{equation}\label{} \nonumber
  J(w)=\frac{w^{T}K_{l}{w}}{w^{T}S_{b}{w}}+\beta{\Arrowvert{w}\Arrowvert_2}=\frac{{\sum\limits_{c \in C} {\sum\limits_{i,j \in c} {{{||{y_i} - {y_j}||}_2^2}} } H_{ij}^c}}{\sum\limits_{i \in c} {{n_i}{{||{u_i} - \bar u||)}_2^2}}}+\beta{\Arrowvert{w}\Arrowvert^2_2}=\frac{{\sum\limits_{c \in C} {\sum\limits_{i,j \in c} {{{||{x_i\cdot w} - {x_j\cdot w}||}_2^2}} } H_{ij}^c}}{\sum\limits_{i \in c} {{n_i}{{||{\bar{x_i}\cdot{w}} - {\bar{x}\cdot{w}}||)}_2^2}}}+\beta{\Arrowvert{w}\Arrowvert^2_2}
\end{equation}
where $\bar{x_i}$ is the mean of the samples belonging class $i$ and $\bar{x}$ is the mean of all samples. To the dimensionality reduction task, the dimension of sample typically exceeds the number of samples. This fact guarantees the dictionary is over-complete, since the projection $w$ plays the role as the coefficient in the collaborative (or sparse) representation while the dimension of samples is the dictionary.
}

\begin{figure}[h]
\setlength{\belowcaptionskip}{-0.3cm}
\centering
\subfigure[]{
\centering
\includegraphics[scale=0.7]{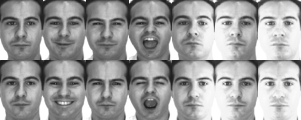}
\label{ar}}
\centering
\subfigure[]{
\centering
\includegraphics[scale=0.32]{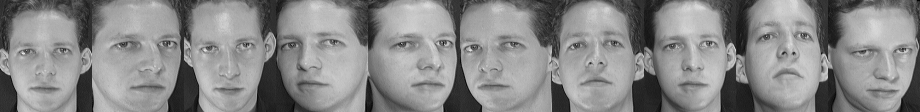}
\label{orl}}
\centering
\subfigure[]{
\includegraphics[scale=0.28]{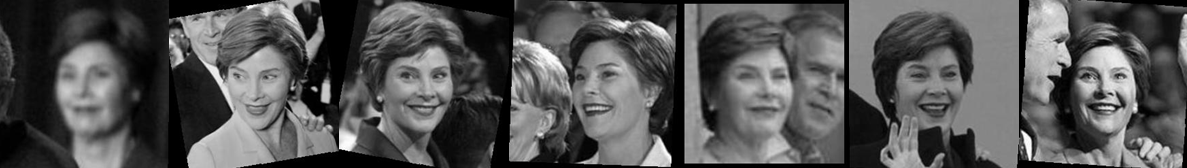}
\label{lfw}
}
\caption{Sample face images from (a) the AR database (b) the ORL database and (c) the LFW-A database}
\label{fig3}
\end{figure}

\section{Experiments}

\subsection{Face Databases}
Three popular face databases including AR \cite{AR}, ORL \cite{ORL} and LFW-A \cite{lfwa} Databases are used to evaluate the recognition performances of the proposed methods.

ORL database contains 400 images from 40 subjects \cite{ORL}. Each subject has ten images acquired at different times. In this database, the subjects' facial expressions and facial details are varying. And the images are also taken with a tolerance for some tilting and rotation. The size of face image on ORL database is 32$\times$32 pixels.

AR database consists of more than 4,000 color images of 126 subjects \cite{AR}. The database characterizes divergence from ideal conditions by incorporating various facial expressions, luminance alterations, and occlusion modes. Following paper \cite{lrc}, a subset contains 1680 images with 120 subjects are constructed in our experiment. The size of face image on AR database is 50$\times$40 pixels.

LFW-A database is an automatically aligned version \cite{lfwa} of LFW (Labeled Faces in the Wild) database which is a very recent database. And it aims at studying the problem of the unconstrained face recognition. This database is considered as one of the most challenging database since it contains 13233 images with great variations in terms of lighting, pose, age, and even image quality. We copped these images to 120$\times$120 pixels around their center and resize these images to 64$\times$64 pixels. On this database, we follow the experimental configuration of \cite{rcr} that uses Local Binary Pattern (LBP) descriptor \cite{lbp} as the baseline image representation.

\subsection {Compared Algorithms}
Nine state-of-the-art face recognition algorithms are used to compare with Class-Scattering Locality Preserving Projections (CSLPP) and Collaborative Discriminant Locality Preserving Projections (CDLPP). Among them, Principal Component analysis (PCA) \cite{pca}, Linear Discriminant Analysis (LDA) \cite{lda}, Locality Preserving Projections (LPP) \cite{lpp}, Discriminant Locality Preserving Projections (DLPP) \cite{dlpp} and Neighbour Preserving Embedding (NPE) \cite{npe} are subspace learning algorithms. Relaxed Collaborative Representation (RCR) \cite{rcr}, Linear Regression Classification (LRC) \cite{lrc}, Sparse Representation Classification (SRC) \cite{sparse}, and Collaborative Representation Classification (CRC) \cite{crc} are not the subspace learning algorithms, but they are four popular face recognition algorithms in recent days. The experiment results of CRC and SRC are directly referenced from the experiment results reported in \cite{rcr} while the experiment results of LRC and RCR are obtained via running the codes by ourselves.

The codes of LDA, PCA, LPP and NPE is downloaded from Prof. Deng Cai's web page \cite{codes}. The code of RCR is provided by Dr. Meng Yang and the code of LRC is provided Mr. Peng Ma.

\begin{table}[h]
\setlength{\belowcaptionskip}{-0.5cm}
\begin{center}
    \begin{tabular}{c c c c c}
    \hline
     \multirow{2}*{Methods}
    &\multicolumn{4}{c}{Recognition Rates $\pm$ Standard Deviation}\\ \cline{2-5}
    & Leave-one-out& 7-fold & 3-fold & 2-fold  \\
    \hline
    PCA \cite{pca} &96.96\%$\pm$2.71\%&93.69\%$\pm$7.29\%&89.17\%$\pm$5.20\%&66.73\%$\pm$0.08\%\\
    LDA \cite{lda} &96.31\%$\pm$4.60\%&96.65\%$\pm$3.38\%&93.04\%$\pm$3.03\%&58.57\%$\pm$1.52\%\\
    NPE \cite{npe} &93.75\%$\pm$6.83\%&92.62\%$\pm$5.15\%&90.83\%$\pm$5.46\%&61.61\%$\pm$0.25\%\\
      LPP \cite{lpp} &93.93\%$\pm$6.57\%&92.56\%$\pm$4.84\% &91.25\%$\pm$4.58\%&61.19\%$\pm$0.34\% \\
      DLPP \cite{dlpp} &95.06\%$\pm$5.46\%&94.29\%$\pm$3.81\% &92.92\%$\pm$4.17\%&65.95\%$\pm$2.69\% \\

      CSLPP &97.44\%$\pm$3.00\%&97.02\%$\pm$2.61\% &94.93\%$\pm$3.09\%&63.45\%$\pm$3.37\%  \\
      CDLPP&\textbf{99.70\%$\pm$0.53\%}&\textbf{99.52\%$\pm$0.85\%}&\textbf{99.31\%$\pm$1.20\%}&69.23\%$\pm$3.11\% \\
 \hdashline
            RCR \cite{rcr} &99.40\%$\pm$0.69\%&99.11\%$\pm$1.03\%& 98.40\%$\pm$1.39\%&\textbf{76.96\%$\pm$1.43\%} \\
      LRC \cite{lrc} &99.58\%$\pm$0.63\%&99.40\%$\pm$0.63\%& 98.47\%$\pm$1.77\%&68.75\%$\pm$0.43\% \\
      \hline
    \end{tabular}

    \caption{Recognition performance comparison (in percents) using AR database}
    \label{arfr}
\end{center}
\end{table}

\begin{table}[h]
\setlength{\belowcaptionskip}{-0.5cm}
\begin{center}
    \begin{tabular}{c c c c c}
    \hline
     \multirow{2}*{Methods}
    &\multicolumn{4}{c}{Recognition Rates $\pm$ Standard Deviation}\\ \cline{2-5}
    & Leave-one-out & 5-fold & 3-fold & 2-fold  \\
    \hline
    PCA \cite{pca} &94.25\%$\pm$3.13\% &91.25\%$\pm$3.19\%&89.72\%$\pm$3.19\%&85.25\%$\pm$0.35\%\\
    LDA \cite{lda} &96.75\%$\pm$3.34\% &96.25\%$\pm$1.98\%&95.83\%$\pm$3.00\%&93.00\%$\pm$0.71\%\\
    NPE \cite{npe} &97.50\%$\pm$3.12\% &94.50\%$\pm$1.90\%&92.22\%$\pm$1.73\%&90.00\%$\pm$4.54\%\\
      LPP \cite{lpp} &98.00\%$\pm$2.58\% &96.75\%$\pm$1.43\%&94.72\%$\pm$3.37\% &90.75\%$\pm$3.89\% \\
      DLPP \cite{dlpp} &98.25\%$\pm$2.06\% &97.25\%$\pm$2.05\%&97.22\%$\pm$2.10\% &93.75\%$\pm$3.18\% \\
      CSLPP &99.00\%$\pm$1.75\% &98.00\%$\pm$1.43\%&97.50\%$\pm$1.44\% &94.50\%$\pm$1.41\% \\
     CDLPP &\textbf{99.00\%$\pm$1.75\%} &\textbf{99.00\%$\pm$1.63\%}&\textbf{98.06\%$\pm$2.10\%} &\textbf{95.75\%$\pm$1.77\% } \\
\hdashline
           RCR \cite{rcr} &98.25\%$\pm$2.06\% &97.5\%$\pm$1.53\%&95.83\%$\pm$1.67\% &93.75\%$\pm$3.89\% \\
      LRC \cite{lrc} &97.50\%$\pm$2.75\% &97.75\%$\pm$2.05\%&96.11\%$\pm$2.92\% &88.75\%$\pm$3.18\% \\
      \hline
    \end{tabular}

    \caption{Recognition performance comparison (in percents) using ORL database}
    \label{orlfr}
\end{center}
\end{table}

\begin{table}[h]
\setlength{\belowcaptionskip}{-0.5cm}
\begin{center}

    \begin{tabular}{c p{4cm}<{\centering} p{4cm}<{\centering}}
    \hline
     \multirow{2}*{LFW-A database}
    &\multicolumn{2}{c}{Top Recognition Rate (Retained Dimensions)}\\ \cline{2-3}
    & subset1 & subset2 \\
    \hline
    PCA \cite{pca}&  35.34\%(529) & 34.25\%(1261)\\
    LDA \cite{lda}& 58.90\%(141)&67.17\%(125)\\
     NPE \cite{npe}& 61.37\%(145)&65.83\%(191)\\
    LPP \cite{lpp}& 58.90\%(145)&65.12\%(169)\\
    DLPP \cite{dlpp}&55.34\%(289) &58.84\%(127) \\
    CSLPP &63.56\%(145) & 71.44\%(127) \\
    CDLPP & \textbf{67.12\%(145)} &\textbf{81.41\%(127)}	\\
\hdashline
   RCR \cite{rcr}&65.75\% &75.17\% \\
    LRC \cite{lrc}&48.49\% &51.76\% \\
SRC \cite{sparse}\footnote{2} &53.00\%& 72.20\% \\
CRC \cite{crc}\footnotemark[\value{footnote}] &54.50\% & 73.00\% \\
  \hline
    \end{tabular}
    \caption{Recognition performance comparison (in percents) using LFW-A database}
    \label{lfwafr}
\end{center}

\end{table}

\subsection{Face Recognition}
Following the conventional subspace learning based face recognition framework, Nearest Neighbour (NN) Classifier is used for classification and the distance metric is the Euclidean distance. With regard to the choice of weighting schemes for the LPP algorithms, we follow the experimental configuration of LPP \cite{lpp} and apply dot-product weighting to construct Laplacian matrices for other LPP algorithms. We use the cross validation scheme to evaluate different algorithms on both AR and ORL databases, since the sample number of each subject is the same on these two databases. The $n$-fold cross validation is defined as follows: the dataset is averagely divided into $n$ parts, $n$-1 parts are used for training and the remainder is used for testing. With regard to LFW-a database, we cannot directly use the cross validation since the subjects of LFW-a database have different sample numbers. Therefore, we follow the experimental way of paper \cite{rcr} and divide the LFW-a database into two subsets. The first subset (147 subjects, 1100 samples) is constructed by the subjects whose sample numbers are ranged from 5 to 10 and the second subset (127 subjects, 2891 samples) is constructed by the subjects whose sample numbers are all over 11. In the experiments, the first five samples of each subject in the first subset will be used for training and the rest samples will be used for testing. Similarly, the first ten samples of each subject in the second subset will be used for training and the remainders are used for testing.

\footnotetext{These results of experiments are directly reference from work \cite{rcr}. However, the experimental configurations of LFW-A database in \cite{rcr} and our works are different.  The baseline feature of \cite{rcr} is the concatenation of four features, including intensity value, low-frequency Fourier feature, Gabor feature and LBP, with a LDA-based discriminative selection while our baseline feature is LBP. According to the performances of LRC and RCR in our work and \cite{rcr}, we deduce that the really recognition accuracies of SRC and CRC in the first subset of LFW-A database under our experimental configuration might be a little bit higher than the ones reported in \cite{rcr} while those accuracies in the second subset of LFW-A database under our experimental configuration might be a little bit lower than the ones reported in \cite{rcr}. }

From the observations of Table \ref{arfr}, Table \ref{orlfr} and Table \ref{lfwafr}, CSLPP outperforms all the compared subspace learning algorithms and CDLPP outperforms all the compared face recognition approaches on all databases. For example, CDLPP obtains absolute improvements around 2\% and 6\% in comparison with the second best face recognition approach in ORL database and the second subset of LFW-A database respectively. The results of face recognition experiments also show that the CDLPP presents a prominent improvement over DLPP. More specifically, the gains of CDLPP over DLPP are around 2\% and around 5\% on ORL and AR databases respectively. On LFW-A database, which is a more challenging database, CDLPP performs even better. The gains of CDLPP over DLPP are 15\% and 23\% in the first subset and the second subset respectively. Moreover, CSLPP also defeats DLPP in all experiments and this verifies that the between-class scatter matrix has a better class scattering ability than the Laplacian matrix of classes.

Recently, the linear regression based face recognition approaches are very popular and generally considered as a more advanced face recognition approach than the conventional Nearest Neighbour Classifier based Subspace learning approach. However, another interesting point learned from our experimental results is that the proposed subspace learning algorithm, CDLPP, consistently outperforms LRC, CRC, SRC and RCR, which are four recent representative algorithms of linear regression based face recognition approach. For instance, CDLPP obtains 6\% , 8\%, 9\% and 30\% more recognition accuracies than RCR, CRC, SRC and LRC respectively in the second subset of LFW-A database. Therefore we believe such phenomenon demonstrates that the conventional subspace learning algorithms may still have the potential to outperform other categories of face recognition algorithms, such linear regression based face recognition approach.

\begin{figure}[h]
\setlength{\belowcaptionskip}{-0.4cm}
\centering
\subfigure[]{
\centering
\includegraphics[scale=0.5]{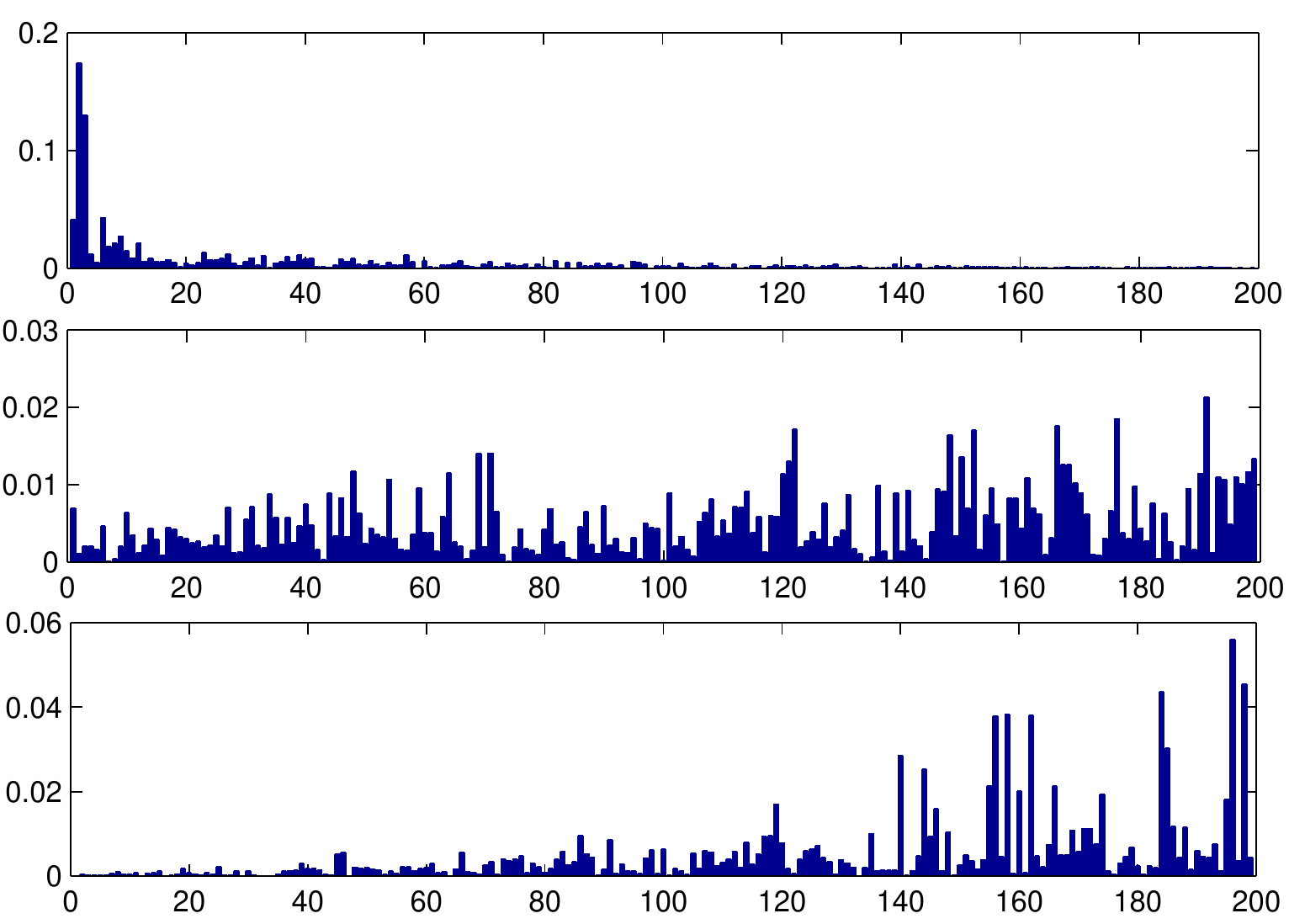}
\label{sparses}}
\centering
\subfigure[]{
\centering
\includegraphics[scale=0.47]{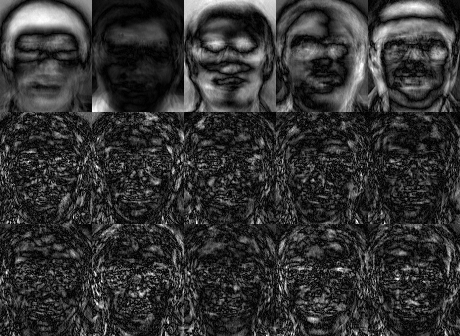}
\label{faces}}
\caption{ (a) the sparseness of the first base of CDLPP, CSLPP and DLPP from top to bottom, (b) the visualizations of the first five bases of CDLPP, CSLPP and DLPP from top to bottom.}
\label{faces}
\end{figure}

In order to show the spareness of the bases of CDLPP and the collaborations between elements of CDLPP base, several experiments are conducted on ORL database. We draw the absolute value of first base of CDLPP, CSLPP and DLPP, $|w|$ , from top to bottom in Figure \ref{sparses}. Clearly, among the three bases, the base of CDLPP is the most sparse one, which verifies the imposed constraint is a relaxed sparse constraint. Moreover, we also visualize the first five bases of CDLPP, CSLPP and DLPP from top to bottom. The brighter part of visualized bases are the elements of base owns a greater magnitude. Comparing with CSLPP and DLPP, such brighter elements of CDLPP always group together to present a facial component. For example, we can clearly find the brightest part is the hair of human in the first visualized base of CDLPP. Such phenomenon verifies that the fact of the dimensions exist collaboration.

\subsection{Dimensionality Reduction }
\begin{figure}[h]
\setlength{\belowcaptionskip}{-0.3cm}
\centering
\subfigure[ ]{
\centering
\includegraphics[scale=0.5]{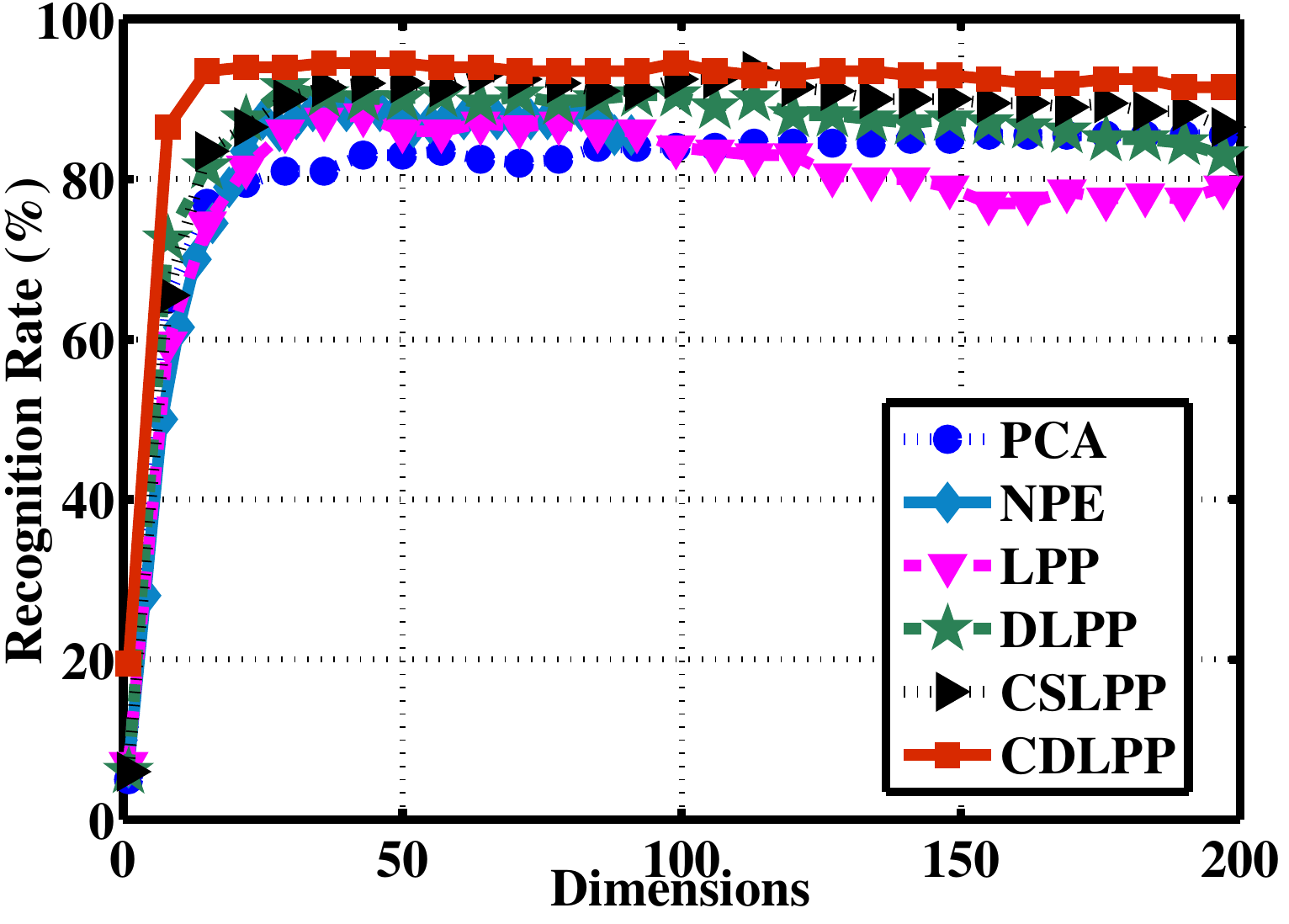}
\label{fig:subfig:a}}
\subfigure[ ]{
\includegraphics[scale=0.5]{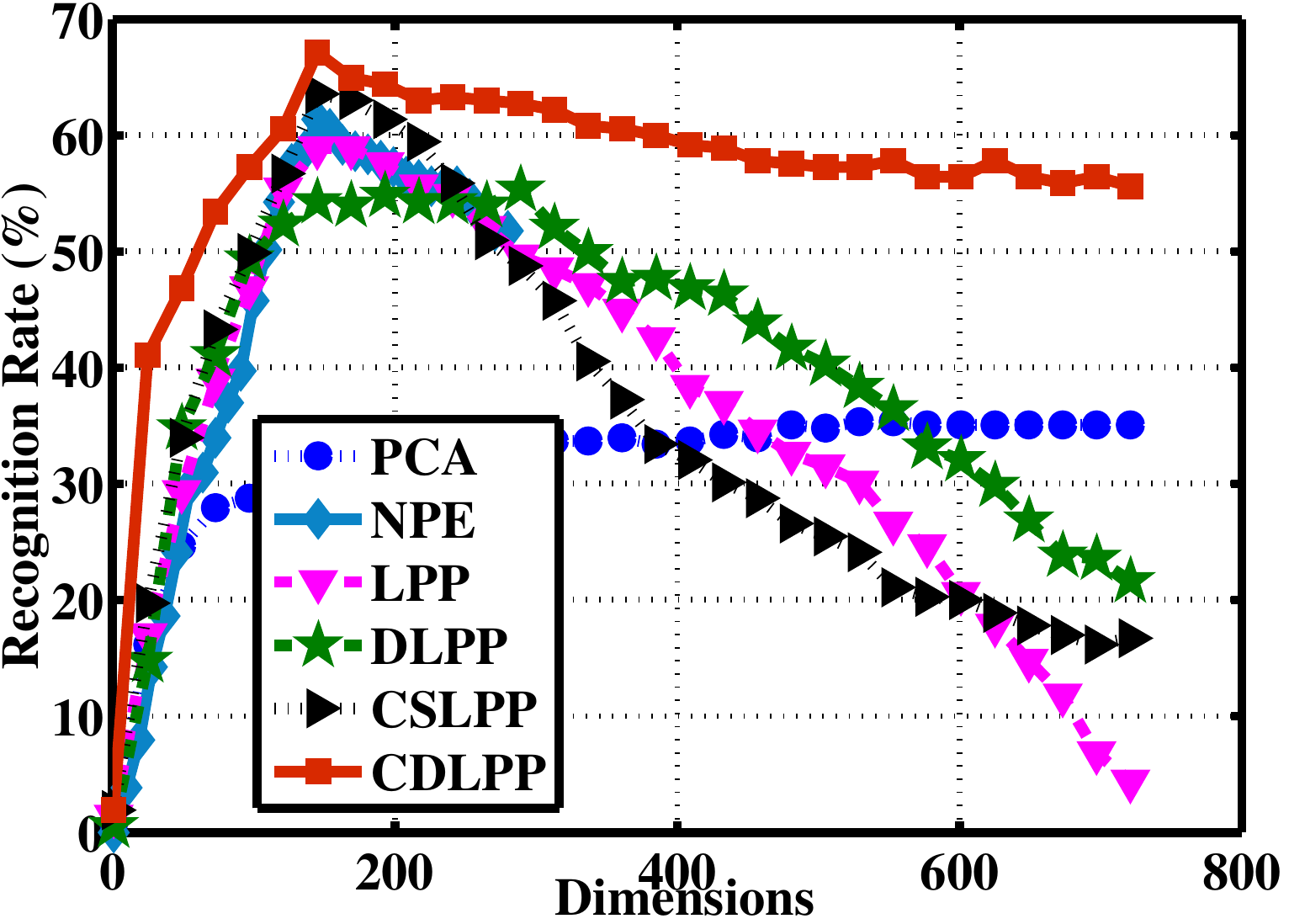}
}
\subfigure[ ]{
\centering
\includegraphics[scale=0.5]{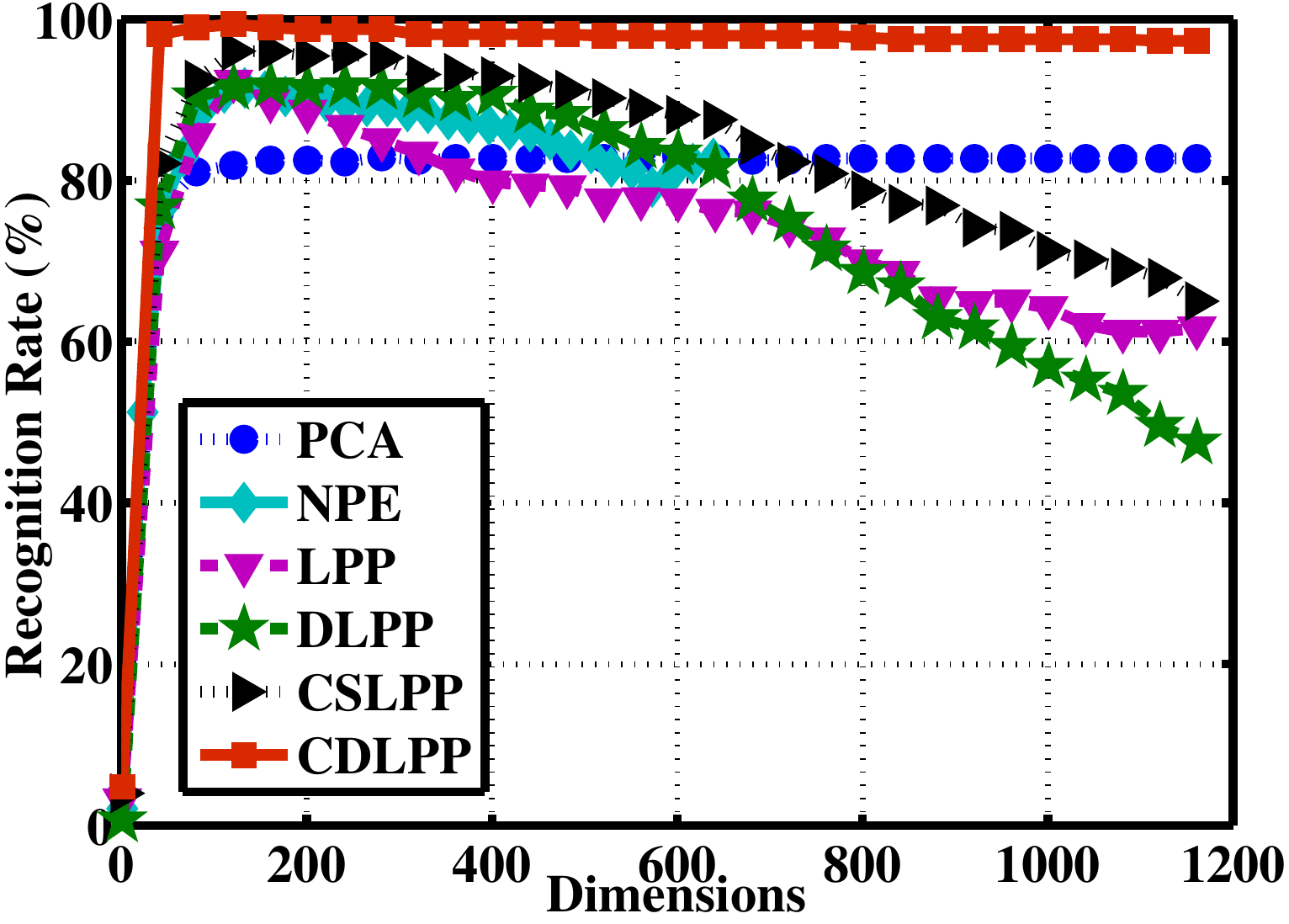}
}
\subfigure[ ]{
\includegraphics[scale=0.5]{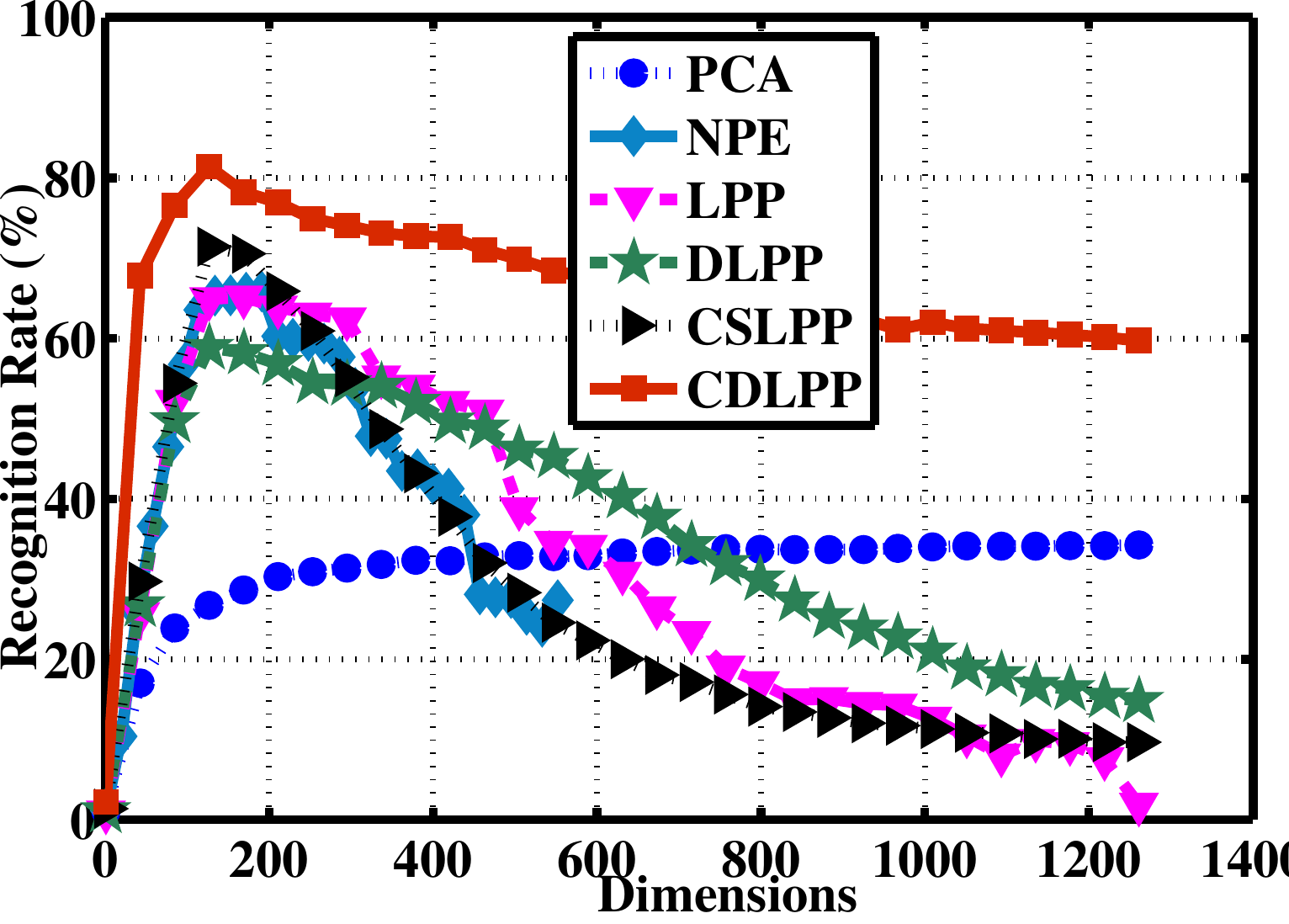}
}
\caption{The recognition rates of different methods corresponding to the dimensions on (a)ORL database (five trains), (b) the first subset of LFW-A database (five trains), (c) AR database (ten trains) and (d) the second subset of LFW-A database (ten trains). \emph{Our methods are the red and black curves}.}
\label{DVR}
\end{figure}

In this section, some experiments are conducted to the dimensionality reduction abilities of different subspace learning algorithms. According to the experimental results in Figure \ref{DVR}, we find that CDLPP consistently outperforms all the subspace learning algorithms cross all the dimensions with a distinct advantage on all databases and the proposed algorithm, CSLPP, also gets a second top recognition accuracy among the subspace learning algorithms. Moreover, CDLPP is a more robust subspace learning algorithm. The recognition accuracy of CDLPP almost not drop down along with the dimension increasing after it reached the top. Actually, this is a very desirable property, since it can facilitate the determination of the optimal number of the retained bases.

\subsection{Training Efficiency}
We examine the training costs of CSLPP and CDLPP, and compare it with LDA, PCA, NPE, LPP and DLPP. The experimental hardware configuration is CPU: 2.2 GHz, RAM: 2G. Table \ref{times} shows the CPU times spent on the training phases by these linear methods using MATLAB. In this experiment, we select five samples of each subject for training. According to the experimental results of Table \ref{times}, CSLPP has a similar training time of the LPP and the training time of CDLPP is the twice of the training time of LPP. Moreover, the training time of all methods on ORL database is almost identical. This is because the size of ORL database is too small and the time of program loading accounts for a great proportion.

\begin{table}[h]
\setlength{\belowcaptionskip}{-0.5cm}
\small
\begin{center}
    \begin{tabular}{c p{2.4cm}<{\centering} p{2.4cm}<{\centering} p{2.4cm}<{\centering} p{2.4cm}<{\centering}}
    \hline
     \multirow{2}*{Methods}
    &\multicolumn{4}{c}{Training Efficiency (Seconds)}\\ \cline{2-5}
    & AR & ORL & LFW-A set1& LFW-A set2 \\
    \hline
    PCA \cite{pca} &2.7768&0.1560&5.3664 &21.7621\\
    LDA \cite{lda} &1.8252&0.1404&3.8064&15.3349\\
    NPE \cite{npe} &3.7752&0.2964&8.6425&35.3498\\
      LPP \cite{lpp} &5.5692&0.3432&10.0777&41.3403\\
      DLPP \cite{dlpp} &8.5489&0.4212&14.6017&63.1336\\
      CSLPP &5.5068&0.2340&10.7641 &45.9267\\
     CDLPP &10.1869&0.3588&19.6717&92.2746\\
      \hline
    \end{tabular}

    \caption{Training times comparison (in seconds) using different databases.}
    \label{times}
\end{center}
\end{table}

\subsection{Parameter Selection of CDLPP}
\begin{figure}[h]
\setlength{\belowcaptionskip}{-0.5cm}
  \centering
  \subfigure[ ]{
  \includegraphics[scale=0.5]{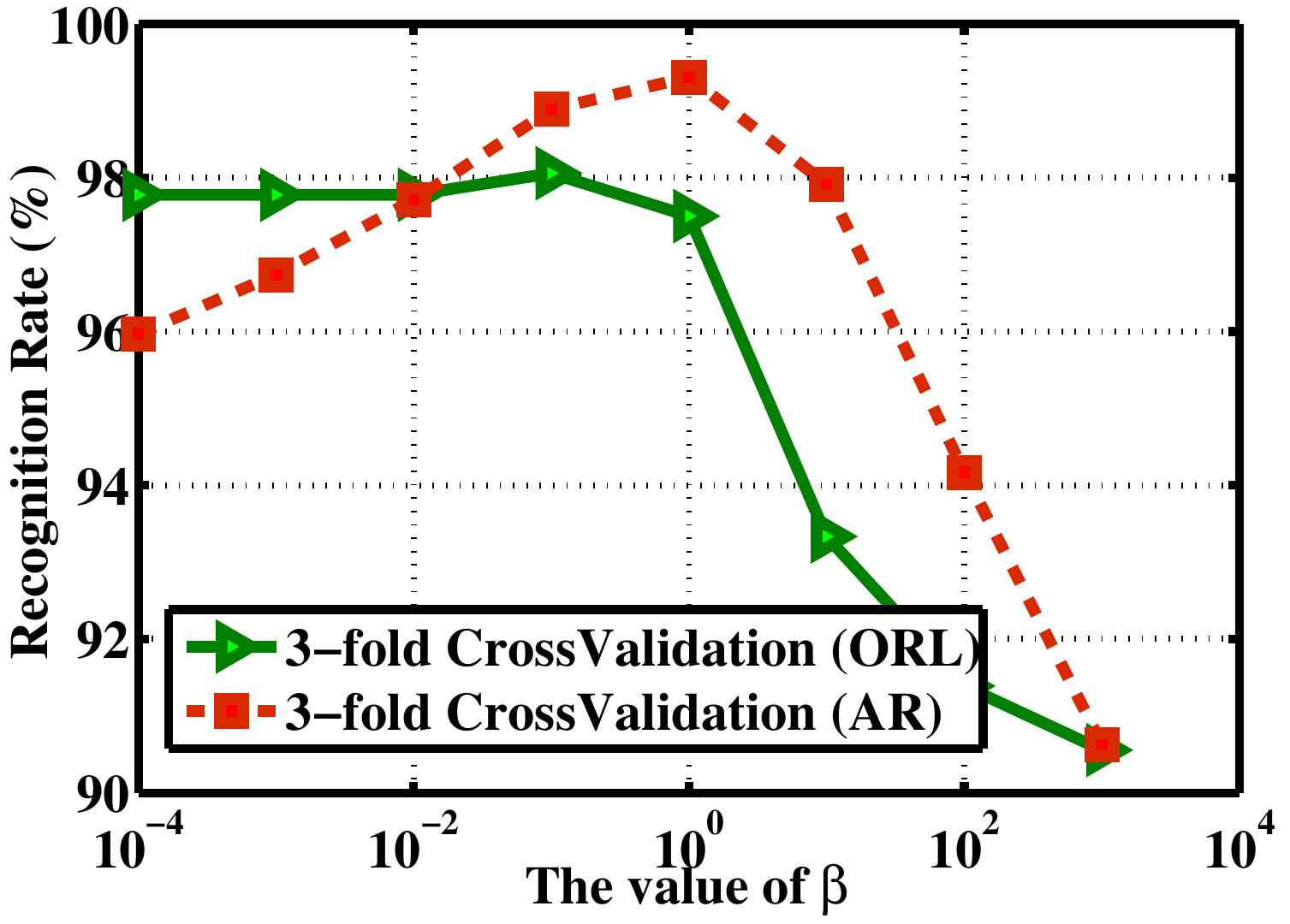}\label{ar_beta}}
    \subfigure[ ]{
  \includegraphics[scale=0.5]{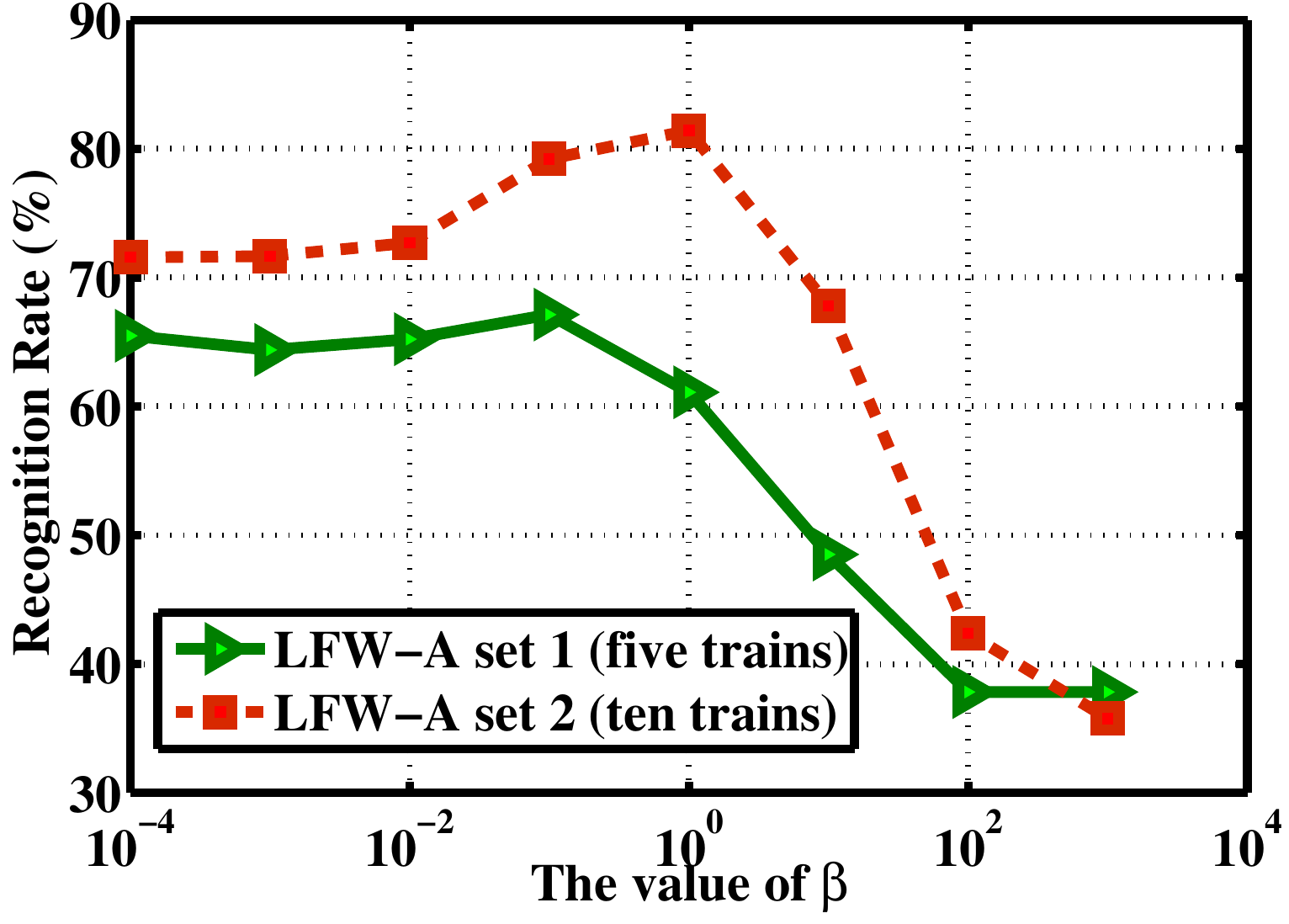}\label{lfw_beta}}
  \caption{The recognition rates under different $\beta$ on (a) AR and ORL databases, and (b) LFW-A database.}
  \label{beta}
\end{figure}

$\beta$ is an important parameter to control the amount of additional collaborations. Figure \ref{beta} depicts the effect of $\beta$ to the recognition performance of CDLPP. The curves plot the relationship between recognition rate and $\beta$ on ORL, AR and LFW-A databases. According to the observations of Figure \ref{beta}, we can know that the recognition accuracies are slowly increasing along with adding more collaborations in the beginning. While, after it reaches the top, the accuracies are decreasing dramatically. This phenomenon verifies that the moderate collaborations of dimension can offer a significant contribution to improve the discriminating power of DLPP while the overmuch collaborations can degrade the model (Equation \ref{l2scslpp}) into the minimizing of the $L_2$-norm of the projections, which is meaningless. Another phenomenon is that a large database seems can be better benefitted by the collaborations of dimensions. This is very desirable for a particular application, since the amount of samples in the particular application is always very large.

According to the results of these experiments, we let $\beta=0.1$ in the experiments using ORL database and the first subset of LFW-A database and let $\beta=1$ in the experiments using AR database and the second subset of LFW-A database.

\section{Conclusion}
In this paper, we present a novel DLPP algorithm name Collaborative Discriminant Locality Preserving projections (CDLPP) and apply it to face recognition. In this algorithm, we use the between-class scatter matrix to replace the original denominator of DLPP to guarantee the global optimum of classes scattering. Motivated by the idea of collaborative representation, a $L_2$-norm constraint is imposed to the projection $w$ as a collaborations constraint for improving the quality of bases. Three popular face databases, including ORL, AR and LFW-A databases, are employed for testing the proposed algorithms. CDLPP outperforms all the compared state-of-the-art face recognition approaches. Our next work may focus on utilizing the collaborations of dimensions to solve the feature selection and image segmentation tasks.
\section*{Acknowledgement}
This work has been supported by the Fundamental Research Funds for the Central Universities (No. CDJXS11181162 and CDJZR12098801) and the authors would thank Dr. Lin Zhong and Dr. Amr bakry for their useful suggestions.
\label{}

\bibliographystyle{unsrt}
\bibliography{mybib}

\end{document}